\documentclass[sigplan, nonacm]{acmart}

\AtBeginDocument{%
  }

\copyrightyear{2026}
\acmYear{2026}
\setcopyright{cc}
\setcctype{by}
\acmConference[GLSVLSI '26]{Great Lakes Symposium on VLSI 2026}{June 22--24, 2026}{Finger Lakes, NY, USA}
\acmBooktitle{Great Lakes Symposium on VLSI 2026 (GLSVLSI '26), June 22--24, 2026, Canandaigua, NY, USA}

\usepackage{graphicx}   
\usepackage{multirow}
\usepackage{makecell}
\usepackage{comment}
\usepackage{enumitem}


\begin{document}

\title[Lightweight SAR Ship Detection via Contrastive Distillation]{Lightweight SAR Ship Detection \\via Contrastive Distillation}

\author{Surendar Devasundaram}
\affiliation{%
  \institution{University of Arizona}
  \department{Department of Electrical and Computer Engineering}
  \city{Tucson}
  \state{Arizona}
  \country{USA}
}
\email{surdev@arizona.edu}

\author{Banafsheh Saber Latibari}
\affiliation{%
  \institution{University of Arizona}
  \department{Department of Electrical and Computer Engineering}
  \city{Tucson}
  \state{Arizona}
  \country{USA}
}
\email{banafsheh@arizona.edu}

\author{Abhijit Mahalanobis}
\affiliation{%
  \institution{University of Arizona}
  \department{Department of Electrical and Computer Engineering}
  \city{Tucson}
  \state{Arizona}
  \country{USA}
}
\email{amahalan@arizona.edu}

\renewcommand{\shortauthors}{Devasundaram et al.}

\begin{abstract}
Deep convolutional and transformer-based detectors achieve strong performance for SAR
ship detection but are often computationally prohibitive for real-time or onboard 
deployment. Lightweight models offer improved efficiency yet struggle to capture the 
complex structural relationships inherent in SAR backscatter. Most existing SAR 
knowledge-distillation approaches rely on feature or logit matching, which enforces localized activation 
similarity while neglecting the geometric relationships among object representations. We propose 
a \textbf{S}tructured \textbf{U}nified \textbf{R}elational knowled\textbf{GE} distillation framework for 
\textbf{SAR Ship} detection \textbf{(SURGE)} that transfers relational geometry from a powerful 
teacher detector to a compact student detector using a contrastive InfoNCE objective in a shared 
projection embedding space. To the best of our knowledge, this work presents the first transformer-based 
SAR ship detector knowledge distillation framework in SAR domain. The framework is architecture-agnostic in the sense that it provides a common region-level distillation interface for two-stage, one-stage and transformer-based detectors without modifying their deployed architectures.
Experiments on the SSDD and HRSID benchmarks demonstrate that the proposed method 
yields substantial improvements for two-stage detectors, achieving up to \textbf{6.2 mAP} and 
\textbf{8.0 AP\(_{75}\)} gains over baseline student and even surpassing teacher performance. 
\end{abstract}



\maketitle

\section{Introduction}
\label{sec:intro}

Synthetic Aperture Radar (SAR) imaging enables robust object detection under adverse 
weather and illumination conditions, making it indispensable for maritime 
surveillance, disaster management, and defense applications. Unlike optical imagery, 
SAR captures backscattering responses from surface structures, providing critical 
geometric and material information even through clouds and darkness \cite{6504845}. 
However, the strong speckle noise, complex scattering mechanisms, and fine structural 
variations of SAR scenes pose unique challenges for deep object detectors 
\cite{zhu2021deep}.

\indent High-capacity convolutional and transformer-based detectors such as Faster R-
CNN \cite{girshick2015fast}, RetinaNet \cite{Lin_2017_ICCV}, and DETR 
\cite{carion2020end} achieve strong detection accuracy, and their adaptation to SAR 
data has yielded high-performing but computationally expensive models 
\cite{li2017ship, gao2022retinanet, feng2023oegr}. Such models are often unsuitable 
for real-time or onboard deployment due to memory and latency constraints. In 
contrast, lightweight detectors improve efficiency but frequently lose discriminative
power, struggling to capture the structural relationships that distinguish targets 
from clutter in SAR backscatter.
Knowledge distillation (KD) has been explored to transfer knowledge from large 
teacher models to compact students. However, \textit{existing SAR-oriented KD methods 
primarily rely on feature-map matching} \cite{min2019gradually, wang2021boosting} or 
\textit{logit alignment} \cite{han2024improving, wang2021sar}, enforcing localized activation 
similarity. These approaches do not explicitly model the relational geometry that 
governs how a teacher organizes object representations in feature space. As a result, 
the student learns local responses rather than the structural reasoning employed by 
the teacher—an important limitation in SAR imagery, where contextual and geometric 
cues are critical.
We address this limitation by proposing a relation-aware knowledge 
distillation framework that transfers object-level relational geometry from a teacher 
detector to a compact student. Our method formulates distillation as a contrastive 
Information Noise-Contrastive Estimation (InfoNCE) objective in a shared projection 
embedding space, where normalized teacher and student region features are compared via
pairwise similarity.We emphasize that the novelty is not a new general-purpose contrastive objective, but a detection-aware formulation for SAR ship detection: teacher predictions from heterogeneous detector families are converted into 
candidate object regions, aligned across teacher and student preprocessing spaces, 
and used to supervise object/region embeddings rather than whole-image 
representations or local activation maps. This enables an architecture-agnostic 
region-level distillation interface for two-stage, one-stage, and 
transformer-based detectors while encouraging the student to preserve the 
teacher's semantic topology instead of merely matching pixel-level or 
activation-level responses. Among the evaluated detector families, the two-stage Faster R-CNN architecture is the strongest choice for SAR ship detection and SURGE provides its most substantial improvement on this family.
Our main contributions are summarized as follows:
\begin{itemize}[leftmargin=*, topsep=2pt, itemsep=1pt, parsep=0pt, partopsep=0pt]
    \item We propose \textbf{SURGE}, a unified relation-aware knowledge distillation framework 
    for SAR ship detection. The novelty lies in converting heterogeneous detector predictions 
    into aligned candidate regions and transferring relational geometry among object-level 
    embeddings, rather than introducing a new general-purpose InfoNCE loss.

    \item We introduce a \textbf{unified region-level distillation interface} applicable to 
    two-stage, one-stage, and transformer-based detectors without modifying their core designs.

    \item To the best of our knowledge, we present the first transformer-based distillation 
    framework for SAR ship detection, demonstrating effective relational transfer without 
    relying on query-level supervision.

    \item Experiments on SSDD and HRSID show substantial gains for lightweight two-stage 
    detectors, with up to \textbf{6.2 mAP} and \textbf{8.0 AP$_{75}$} improvement while 
    achieving over \textbf{50\% parameter reduction}.
\end{itemize}

\section{Related Works}
\label{sec:format}

Recent work in SAR ship detection has primarily focused on adapting deep object detectors
originally developed for optical imagery~\cite{lang2025recent}. Two-stage detectors such as
Faster R-CNN\cite{li2017ship} and one-stage architectures, including RetinaNet \cite{gao2022retinanet} have been extended to SAR data
using multiscale feature learning and task-specific preprocessing to mitigate speckle noise
and complex scattering effects. More recently, transformer-based detectors inspired by DETR
have been explored to leverage global context modeling in SAR imagery~\cite{yin2025ship}.
While these approaches achieve strong detection accuracy, their reliance on deep backbones
and dense feature representations results in high computational complexity, limiting their
suitability for real-time or onboard SAR systems.
Knowledge distillation has been investigated to reduce the computational cost of deep models
in SAR recognition and detection tasks~\cite{min2019gradually, wang2021boosting}. Existing
SAR-oriented distillation methods predominantly rely on feature-map matching or output-level
logit alignment~\cite{wang2024m, yu2023multilevel, yang2022sar}. 
Representative approaches transfer intermediate convolutional features using attention mechanisms, 
multiscale supervision, or privileged information~\cite{lee2021privileged}, and some combine 
distillation with network pruning or lightweight architectural design. Despite their effectiveness, 
these methods primarily enforce localized activation similarity and do not explicitly capture the
relational structure among object representations, with most studies focusing on target
recognition or classification rather than architecture-agnostic SAR ship detection.
Relation-aware KD has been proposed in
general computer vision to preserve structural information in representation spaces
\cite{park2019relational, tung2019similarity}. By transferring pairwise similarities or
distances, these methods enable students to learn the geometric organization of teacher
embeddings. More recently, contrastive objectives such as InfoNCE have been incorporated into
distillation frameworks~\cite{tian2019contrastive, wu2018unsupervised}, improving robustness
and generalization in optical image classification and detection. However, to our knowledge,
relation-aware contrastive distillation has not been systematically studied for SAR ship
detection across heterogeneous detector architectures, motivating the unified region-level
framework proposed in this work.

\begin{figure*}[!t]
    \centering
    \resizebox{0.95\textwidth}{\height}{%
        \includegraphics[width=0.8\textwidth]{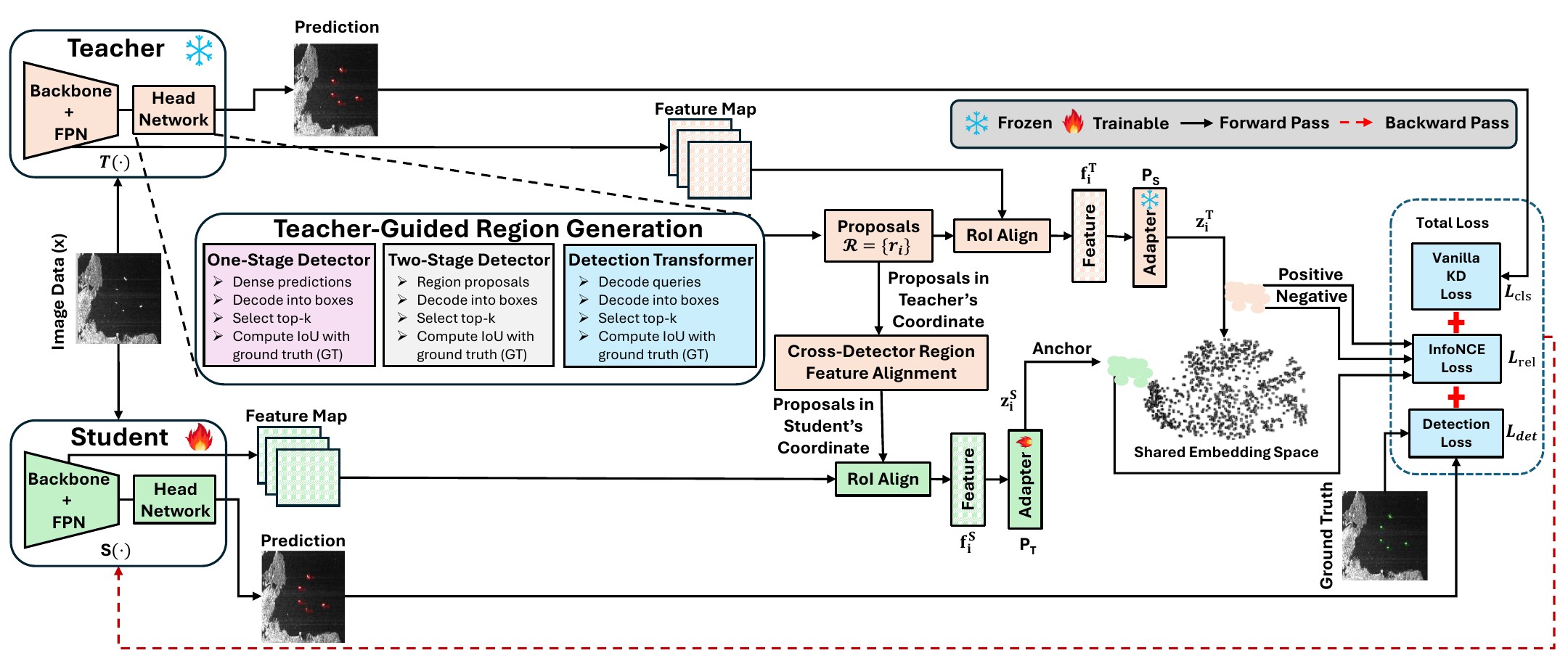}
    }
    \vspace{-5pt}
    \caption{Overview of SURGE framework for SAR ship detection. Given an input SAR image $x$, a frozen teacher detector $T(\cdot)$ and a trainable student detector $S(\cdot)$ extract feature maps using a backbone and FPN. Teacher predictions are decoded into candidate regions $\mathcal{R}=\{r_i\}$ (region proposals for two-stage detectors, dense predictions for one-stage detectors or decoder-predicted bounding boxes in DETR), followed by top-$K$ selection and IoU-based filtering. Teacher regions are mapped to the student image space to ensure spatial alignment, and RoI features $\mathbf{f}_i^T$ and $\mathbf{f}_i^S$ are extracted via RoIAlign and projected using heads $P_T$ and $P_S$ to obtain normalized embeddings $\mathbf{z}_i^T$ and $\mathbf{z}_i^S$. The student is trained by jointly optimizing the detection loss $\mathcal{L}_{\mathrm{det}}$, vanilla distillation losses, and a relation-aware contrastive loss $\mathcal{L}_{\mathrm{rel}}$ based on an InfoNCE objective, encouraging preservation of the teacher’s relational geometry across detector architectures.}
    \label{overview}
\end{figure*}

\vspace{-10pt}
\section{Methodology}
\label{sec:method}

\begin{figure}
    \centering
    \includegraphics[width=0.9\linewidth]{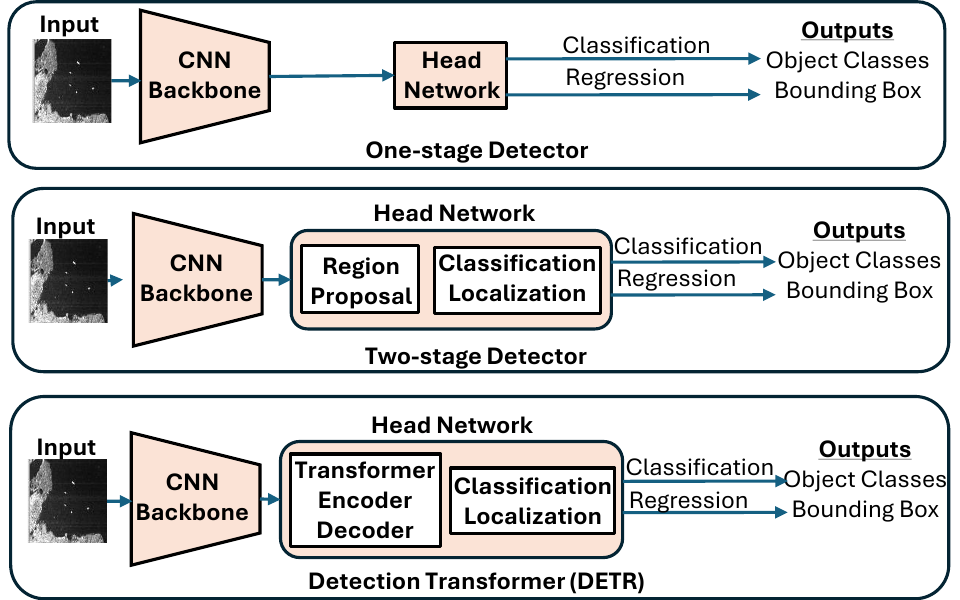}
    \caption{Head network designs for different object detection paradigms.
    \textbf{Top: One-stage detectors} directly predict class scores and bounding boxes from 
    dense backbone feature maps.
    \textbf{Middle: Two-stage detectors} generate region proposals via an RPN, followed by per-proposal
    classification and localization heads.
    \textbf{Bottom: Detection transformers} process CNN features with a transformer encoder–
    decoder, where object queries are decoded to jointly predict class labels and bounding 
    boxes.}
    \label{headnetwork}
\end{figure}

Our SURGE framework is depicted in Fig.~\ref{overview}, where the 
teacher and student detectors process the same input SAR image to extract multi-scale 
feature maps. High-confidence predictions from the frozen teacher are decoded into 
candidate regions, which are spatially aligned with the student and used to extract 
corresponding region features via RoIAlign. These aligned region features are projected
into a shared embedding space and supervised using a contrastive objective, while the 
student is simultaneously optimized with the standard detection loss. Despite 
architectural differences as shown in Fig.~\ref{headnetwork}, all detectors ultimately 
produce classification and localization outputs, which enables a unified region-level 
distillation strategy that preserves the geometric structure of the teacher's 
representation space.

\textbf{Teacher-Guided Region Generation:}
Given an input image $x$, the teacher detector produces a set of candidate regions
$\mathcal{R}=\{r_i\}_{i=1}^{N}$, each associated with a semantic label $c_i$ and a localization
quality score $\text{IoU}_i$ computed with respect to ground truth. These regions serve as
anchors for object-level distillation. For two-stage detectors, regions are obtained directly from
the teacher’s region proposal network (RPN). For one-stage detectors, high-confidence predictions 
are selected from dense outputs, decoded into bounding boxes, and treated as pseudo-proposals. For 
detection transformers, high-confidence bounding boxes predicted by the teacher DETR decoder are used
as candidate regions. In all cases, the resulting regions are used to extract region-level
features via RoIAlign, providing a unified interface across detector paradigms.

\textbf{Cross-Detector Feature Alignment:}
Teacher-derived regions are first mapped back to the ground-truth image coordinate system and
then transformed into the student image space to account for architectural differences in
preprocessing.

\textbf{Relation-Aware Distillation Objective:}
Let $f_i^T$ and $f_i^S$ denote the teacher and student region features corresponding to region
$r_i$. Region features are globally pooled and projected into a shared embedding space:
\begin{equation}
z_i^T = \frac{P_T(\mathrm{GAP}(f_i^T))}{\|P_T(\mathrm{GAP}(f_i^T))\|}, \quad
z_i^S = \frac{P_S(\mathrm{GAP}(f_i^S))}{\|P_S(\mathrm{GAP}(f_i^S))\|},
\end{equation}
where $\mathrm{GAP}(\cdot)$ denotes global average pooling, $P_T$ and $P_S$ are lightweight
projection heads, and embeddings are L2-normalized.
For each student embedding $z_i^S$, positive and negative sets (In all experiments, we use $\tau_{\text{pos}}=0.5$ and $\tau_{\text{neg}}=0.3$) are defined using teacher-derived
semantic and localization cues:
\begin{align}
\mathcal{P}(i) &= \{ j \mid c_j = c_i,\; \text{IoU}_j \ge \tau_{\text{pos}} \}, \\
\mathcal{N}(i) &= \{ j \mid \text{IoU}_j \le \tau_{\text{neg}} \}.
\end{align}
A student-anchored InfoNCE objective is used to preserve the relational geometry of the teacher
representation:
\begin{equation}
\mathcal{L}_{\text{rel}} =
- \frac{1}{|\mathcal{A}|}
\sum_{i \in \mathcal{A}}
\log
\frac{
\sum_{j \in \mathcal{P}(i)} \exp(\mathrm{sim}(i,j))
}{
\sum_{j \in \mathcal{P}(i)\cup\mathcal{N}(i)} \exp(\mathrm{sim}(i,j))
},
\end{equation}
where $\mathcal{A}$ denotes anchors with at least one valid positive, and $\mathrm{sim}(i,j)$ is the temperature-scaled cosine similarity between the
student anchor embedding and the teacher region embedding. To stabilize training, the
number of positives is capped and hard negative mining is applied. The use of InfoNCE is motivated by the multi-positive/multi-negative structure of the proposed supervision. Unlike one-to-one feature regression, the objective encourages a student region embedding to preserve the relative neighborhood structure induced by teacher-derived semantic and localization cues. Thus, the goal is relative separation and relational geometry preservation rather than direct activation matching.
In addition to relational distillation, standard output-level distillation is applied to
classification and bounding-box predictions for convolutional detectors. For two-stage
detectors, this supervision is applied at the RoI head, while for one-stage detectors it is
applied to dense prediction heads. For detection transformers, output-level distillation is
not employed, as the unordered set-based predictions lead to unstable teacher–student
correspondence and were observed to degrade performance. Consequently, DETR models rely
exclusively on region-level relational distillation.
The overall training objective is defined in eq where $\mathcal{L}_{\text{det}}$ is the standard detection loss and $\lambda$ terms balance the
contributions of relational, classification, and localization distillation.

{\setlength{\abovedisplayskip}{2pt}
 \setlength{\belowdisplayskip}{2pt}
 \setlength{\abovedisplayshortskip}{1pt}
 \setlength{\belowdisplayshortskip}{1pt}
\begin{equation}
\mathcal{L} =
\mathcal{L}_{\mathrm{det}}
+\lambda_c \mathcal{L}_{\mathrm{rel}}
+\lambda_{\mathrm{cls}}\mathcal{L}_{\mathrm{cls}}^{\mathrm{KD}}
+\lambda_{\mathrm{box}}\mathcal{L}_{\mathrm{box}}^{\mathrm{KD}} .
\end{equation}}

\enlargethispage{3\baselineskip}
\section{Experimental Setup}
\label{sec:exp}

We evaluate the proposed framework on two public SAR ship detection benchmarks:
the \textbf{SSDD} and the \textbf{HRSID}. SSDD images are resized to $512 \times 512$, while HRSID images 
are resized to $800 \times 800$. Official train/test splits are used in all experiments. Performance is 
measured using COCO-style mean Average Precision (mAP), along with AP$_{50}$ and AP$_{75}$.
Experiments are conducted on two-stage, one-stage, and transformer-based detectors.
For two-stage detection, Faster R-CNN with a ResNet-101 backbone is used as the teacher and
ResNet-18 as the student. For one-stage detection, RetinaNet with ResNet-101 (teacher) and
ResNet-18 (student) backbones is employed. For transformer-based detection, DETR with a
ResNet-101 teacher and a ResNet-50 student is used. All models follow their standard
architectural definitions, and distillation is applied externally without modifying detector
designs. Teacher and student networks are initialized from ImageNet-pretrained weights, with
teacher parameters frozen during distillation.
Relation-aware KD is implemented using a student-anchored InfoNCE loss
defined over region-level features. Positive and negative pairs are selected based on
teacher-derived semantic labels and IoU thresholds, with temperature scaling, capped positives,
and hard negative mining applied for stability. For convolutional detectors, standard
logit-level and bounding-box distillation losses are additionally used, while for DETR,
we do not apply direct query-level supervision (output-level distillation), where teacher and student decoder queries or output slots are matched one-to-one. DETR predictions are unordered set outputs, and direct teacher-student alignment at the decoder-query level can produce unstable correspondences. Instead, teacher-decoded boxes are treated as region anchors, mapped into student image space, and used for RoI-based relational distillation. Consequently, DETR models rely exclusively on region-level relational distillation in our framework.
All models are implemented in PyTorch. Faster R-CNN models are trained using SGD with momentum
0.9 and weight decay $1\times10^{-4}$, while RetinaNet and DETR models use AdamW with the same
weight decay. Convolutional detectors are trained for 50 epochs with a multi-step learning
rate schedule (decay at epochs 15 and 30), while DETR models are finetuned for 300 epochs to
account for slower convergence. Batch size is set to 4 across all experiments. All training is
performed on a single NVIDIA L40S GPU.

\begin{table*}[!t]
\centering
\small
\caption{Performance comparison of CNN-based and transformer-based detectors on SSDD and HRSID.
Results are mean $\pm$ std across 6 runs (1 \& 2 stage) and across 2 runs (Transformer).
$\Delta$ denotes change relative to the non-distilled student (mean values). 
Bold indicates the best student performance per detector.}
\label{tab:main_results}
\setlength{\tabcolsep}{3pt}
\begin{tabular*}{\textwidth}{@{\extracolsep{\fill}}llc ccc ccc@{}}
\toprule
\multirow{2}{*}{Detector} & \multirow{2}{*}{Model} & \multirow{2}{*}{Params (M)} 
& \multicolumn{3}{c}{SSDD} & \multicolumn{3}{c}{HRSID} \\
\cmidrule(lr){4-6} \cmidrule(lr){7-9}
& & & mAP & AP$_{50}$ & AP$_{75}$ & mAP & AP$_{50}$ & AP$_{75}$ \\
\midrule

\multirow{6}{*}{\makecell[l]{Two-Stage\\(Faster R-CNN)}}
& R101 (Teacher) & 63.8
& 66.70$\pm$0.33 & 93.50$\pm$0.80 & 82.26$\pm$0.80
& 62.09$\pm$0.11 & 86.36$\pm$0.07 & 70.76$\pm$0.33 \\

& R18 (Student) & 31.2
& 62.87$\pm$0.49 & 91.08$\pm$0.75 & 74.64$\pm$1.28
& 59.93$\pm$0.19 & 84.65$\pm$0.07 & 69.01$\pm$0.34 \\

& R18 + Vanilla KD & 31.2
& 67.76$\pm$0.66 & 93.81$\pm$0.87 & 81.78$\pm$1.12
& \textbf{65.98}$\pm$0.10 & 87.06$\pm$0.33 & \textbf{77.21}$\pm$0.43 \\

& R18 + Proposed RKD & 31.2
& \textbf{68.03}$\pm$1.35 & \textbf{94.13}$\pm$0.97 & \textbf{82.35}$\pm$1.63
& 65.96$\pm$0.16 & \textbf{88.10}$\pm$0.51 & 76.72$\pm$0.40 \\

& $\Delta$ Student $\rightarrow$ RKD & --
& +5.16 & +3.05 & +7.71
& +6.03 & +3.45 & +7.71 \\

\midrule

\multirow{5}{*}{\makecell[l]{One-Stage\\(RetinaNet)}}
& R101 (Teacher) & 56.6
& 60.99$\pm$0.81 & 92.61$\pm$1.25 & 72.11$\pm$1.19
& 62.68$\pm$0.27 & 88.91$\pm$0.28 & 71.39$\pm$0.35 \\

& R18 (Student) & 23.5
& \textbf{61.12}$\pm$0.72 & 93.73$\pm$1.08 & 70.72$\pm$1.62
& 62.12$\pm$0.28 & 89.03$\pm$0.36 & 71.02$\pm$0.44 \\

& R18 + Vanilla KD & 23.5
& 60.74$\pm$0.36 & \textbf{94.10}$\pm$0.81 & \textbf{70.99}$\pm$0.76
& \textbf{63.21}$\pm$0.50 & \textbf{89.47}$\pm$0.65 & \textbf{73.01}$\pm$0.58 \\

& R18 + Proposed RKD & 23.5
& 60.33$\pm$0.81 & 93.67$\pm$0.52 & 70.65$\pm$1.48
& 62.77$\pm$0.33 & 89.36$\pm$0.53 & 72.12$\pm$0.40 \\

& $\Delta$ Student $\rightarrow$ RKD & --
& $-$0.79 & $-$0.06 & $-$0.07
& +0.65 & +0.33 & +1.10 \\

\midrule

\multirow{4}{*}{\makecell[l]{Transformer\\(DETR)}}
& R101 (Teacher) & 60.3
& 59.33$\pm$1.62 & 90.85$\pm$2.58 & 69.61$\pm$1.84
& 50.49$\pm$0.10 & 74.20$\pm$0.86 & 58.16$\pm$0.65 \\

& R50 (Student) & 41.3
& 59.37$\pm$0.00 & 91.47$\pm$0.73 & 68.94$\pm$0.36
& \textbf{48.77}$\pm$0.12 & 72.87$\pm$0.18 & \textbf{56.14}$\pm$0.59 \\

& R50 + Proposed RKD & 41.3
& \textbf{59.70}$\pm$0.57 & \textbf{91.59}$\pm$0.46 & \textbf{69.02}$\pm$0.44
& 48.60$\pm$0.19 & \textbf{73.04}$\pm$0.15 & 55.95$\pm$0.10 \\

& $\Delta$ Student $\rightarrow$ RKD & --
& +0.33 & +0.12 & +0.08
& $-$0.17 & +0.17 & $-$0.19 \\

\bottomrule
\end{tabular*}
\end{table*}

\section{Results}
\label{sec:majhead}

Table~\ref{tab:main_results} summarizes the performance of teacher, 
student, and distilled models across two-stage, one-stage, and transformer-based detectors on the 
SSDD and HRSID datasets, together with their parameter counts. The comparison reveals that Faster R-CNN family achieves the strongest overall performance for SAR ship detection, and SURGE yields its largest gains on this best-performing detector family.
All convolution-based detector results are 
reported using COCO-style metrics and averaged over six independent runs while DETR's is averaged over 
two independent runs. 
For comparison, we implemented a baseline KD\cite{chen2017learning} based on 
standard object detection distillation from the RGB domain, applying output-level teacher-student 
supervision.

While several SAR-specific KD methods have been proposed \cite{min2019gradually, 
wang2021boosting, wang2024m}, their implementations are not publicly available and largely rely on 
localized feature or logit alignment. The adopted baseline thus serves as a fair and reproducible 
reference for evaluation.
The proposed relation-aware distillation yields the most significant improvements for
two-stage detectors. On SSDD, the Faster R-CNN student with a ResNet-18 backbone improves from
62.87 mAP to 68.03 mAP, corresponding to a gain of +5.16 mAP over the non-distilled student and
+0.27 mAP over baseline distillation. Notably, the distilled student surpasses the ResNet-101
teacher by +1.33 mAP while using approximately 51\% fewer parameters, demonstrating that
relational geometry transfer can compensate for reduced model capacity. Similar trends are observed
on HRSID, where the proposed method improves the student by +6.03 mAP and +7.71 AP$_{75}$, 
highlighting substantial gains in localization accuracy. While baseline distillation already improves 
performance over the student, the proposed relation-aware objective consistently provides larger 
gains, particularly at higher IoU thresholds. These results indicate that preserving inter-region 
relationships is especially beneficial for two-stage detectors, which explicitly model object-level 
representations through region proposals.
On SSDD, the proposed method for one-stage RetinaNet model yields comparable AP$_{75}$ performance but 
slightly lower mAP than both the baseline student and conventional distillation; however, this 
difference falls within the margin of statistical variability (the student model achieves
$61.12 \pm 0.72$ mAP (95\% CI: $[60.36, 61.88]$), vanilla knowledge distillation achieves
$60.74 \pm 0.36$ mAP (95\% CI: $[60.36, 61.11]$), and the proposed method achieves
$60.33 \pm 0.81$ mAP (95\% CI: $[59.48, 61.18]$).
The substantial overlap of confidence intervals indicates that the observed performance
differences are not statistically significant.) On HRSID, however, the 
proposed method improves the student by +0.65 mAP and +1.10 AP$_{75}$, indicating that 
relational distillation can still provide benefits in more complex scenes with higher object 
density. Compared to two-stage detectors, the smaller gains for one-stage models can be 
attributed to the absence of explicit region proposals and the reliance on dense 
anchor-based predictions, which may restrict the expressiveness of region-level relational 
supervision.
For transformer-based detectors, we evaluate DETR using a region-based
distillation strategy derived from teacher-predicted bounding boxes. Although the absolute gains
are smaller, the proposed method consistently improves the student model, achieving +0.33 mAP
over the baseline student and marginal improvements over the teacher. A conventional output-level 
distillation baseline is not reported for DETR, as its unordered set-based predictions lack a stable 
teacher-student conrrespondence and led to degraded performance in our preliminary experiments. The 
results confirm that relational knowledge can be effectively transferred to transformer-based 
detectors through a unified RoI-based interface, even when query-level supervision is not explicitly 
used. 

\textbf{Loss Ablation:}
Table~\ref{tab:loss_ablation} presents an ablation study analyzing the contribution of
different loss components. Vanilla KD alone yields substantial gains over detection-only
training, while relational distillation alone provides limited improvement. The strongest
performance is achieved when relational distillation is combined with vanilla KD, resulting
in a +5.16 mAP and +7.71 AP$_{75}$ improvement over the detection-only baseline. These results
demonstrate that relational supervision is most effective when complementing conventional
teacher--student distillation rather than replacing it.
\begin{table}[!t]
\centering
\small
\caption{Loss ablation on \textbf{Faster R-CNN R18 (SSDD)}.
Mean$\pm$std over $n=6$ runs.}
\label{tab:loss_ablation}
\setlength{\tabcolsep}{3.5pt}
\begin{tabular}{lccc}
\toprule
\textbf{Losses Used} & \textbf{mAP} & \textbf{AP$_{50}$} & \textbf{AP$_{75}$} \\
\midrule
Det only                            
& 62.87$\pm$0.49 & 91.08$\pm$0.75 & 74.64$\pm$1.28 \\

Det + KD                            
& 67.76$\pm$0.66 & 93.81$\pm$0.87 & 81.78$\pm$1.12 \\

Det + Rel                           
& 62.77$\pm$0.33 & 89.36$\pm$0.53 & 72.12$\pm$0.40 \\

Det + KD + Rel (Ours)               
& \textbf{68.03}$\pm$1.35 
& \textbf{94.13}$\pm$0.97 
& \textbf{82.35}$\pm$1.63 \\

\bottomrule
\end{tabular}
\end{table}

\textbf{Computational Efficiency:} Since the proposed method is used only during training, it does not modify the deployed student detector. Therefore, computational cost is identical for the baseline student, vanilla KD student, and SURGE student within the same detector family. SURGE adds training-only overhead from teacher inference, region decoding, RoIAlign, and projection heads, but leaves deployment-time latency and memory unchanged. We report GFLOPs, single-image inference latency, and peak memory usage to quantify computational efficiency in table \ref{tab:efficiency_ssdd}.

\textbf{SOTA Comparison:} Table~\ref{tab:mgd_comparison_ssdd} compares SURGE with the Masked generative distillation (MGD)~\cite{yang2022masked} baseline on SSDD. SURGE improves over MGD across all three detector families, with gains of 
+5.23, +1.63, and +22.34 mAP for Faster R-CNN, RetinaNet, and DETR, respectively. Unlike MGD, which distills knowledge by reconstructing teacher feature maps from randomly masked student features, SURGE performs teacher-guided region-level relational distillation. The proposed method transfers the relative geometry among object-region embeddings rather than enforcing dense feature-map recovery, making it better aligned with SAR ship detection and heterogeneous detector families.

\begin{table}[!t]
\centering
\caption{\textbf{Computational efficiency on SSDD.}}
\vspace{-10pt}
\label{tab:efficiency_ssdd}
\scriptsize
\setlength{\tabcolsep}{3pt}
\resizebox{\columnwidth}{!}{
\begin{tabular}{llccc}
\toprule
\textbf{Detector} & \textbf{Model} & \textbf{GFLOPs} & \textbf{Latency (ms)} & \textbf{Memory (MB)} \\
\midrule
Faster R-CNN & R101 Teacher & 68.64 & 21.02 & 543.76 \\
Faster R-CNN & R18 Student / Ours & 49.77 & 10.53 & 362.52 \\
\midrule
RetinaNet & R101 Teacher & 160.09 & 23.71 & 496.87 \\
RetinaNet & R18 Student / Ours & 127.31 & 13.32 & 316.72 \\
\midrule
DETR & R101 Teacher & 44.36 & 22.50 & 315.63 \\
DETR & R50 Student / Ours & 24.97 & 15.80 & 243.61 \\
\bottomrule
\end{tabular}}

\end{table}

\begin{table}[!t]
\centering
\caption{\textbf{MGD~\cite{yang2022masked} comparison on SSDD.}}
\vspace{-10pt}
\label{tab:mgd_comparison_ssdd}
\scriptsize
\setlength{\tabcolsep}{3pt}
\begin{tabular*}{\columnwidth}{@{\extracolsep{\fill}}lcc@{}}
\toprule
\textbf{Detector} & \textbf{MGD}~\cite{yang2022masked} (mAP) & \textbf{SURGE} (mAP) \\
\midrule
Faster R-CNN R18 & 62.80 & 68.03 (+5.23) \\
RetinaNet R18    & 58.70 & 60.33 (+1.63) \\
DETR R50         & 37.36 & 59.70 (+22.34) \\
\bottomrule
\end{tabular*}
\end{table}

\textbf{Discussion:}
Overall, the results indicate that SURGE is particularly effective for detectors that rely on 
explicit region-level representations, such as two-stage architectures. By transferring relational 
geometry rather than enforcing local feature similarity, the proposed method enables compact student 
models to achieve near-teacher or teacher-surpassing performance with substantial reductions in model 
size. While gains for one-stage and transformer-based detectors are comparatively smaller, the consistent 
improvements across architectures highlight the generality of the proposed framework and its potential 
for efficient SAR ship detection under resource constraints. 

\vspace{-9pt}
\section{Conclusion and Future Direction} 
\label{sec:conclusion} 
We introduced the \textbf{SURGE}, the first relation-aware knowledge distillation 
framework for SAR ship detection that explicitly transfers the relational geometry of 
object representations from a high-capacity teacher detector to a lightweight student. 
By formulating distillation as a \emph{student- anchored contrastive learning} problem 
in a shared embedding space, the proposed method goes beyond conventional feature or 
logit matching and enables the student to preserve the structural organization of the 
teacher’s representation space. The framework is \emph{architecture-agnostic} and 
operates through a unified region-level distillation interface, allowing seamless 
application to two-stage, one-stage, and transformer-based detectors without modifying 
their core architectures. To the best of our knowledge, this work is also the first to 
demonstrate effective knowledge distillation for transformer-based SAR ship detectors, 
enabling relational transfer without relying on query-level supervision. 
Extensive experiments on the SSDD and HRSID benchmarks demonstrate that the proposed 
approach is particularly effective for two-stage detectors, where it achieves up to 
\textbf{6.2 mAP} and \textbf{8.0 AP$_{75}$} improvements over non-distilled students 
while reducing model size by more than \textbf{50\%}. In several cases, the distilled 
student matches or surpasses teacher performance, highlighting the importance of 
transferring relational geometry for SAR imagery. For one-stage and transformer-based 
detectors, the method yields consistent but more modest gains, confirming its generality 
while revealing the varying impact of relational supervision across detector paradigms. 
Overall, this work establishes relation-aware distillation as a principled and scalable 
strategy for efficient SAR ship detection under resource constraints. Future work will 
explore extensions to multi- class SAR detection, stronger transformer architectures, 
and cross-modal distillation between SAR and optical domains.

\vspace{-5pt}
\bibliographystyle{ACM-Reference-Format}
\bibliography{refs}

\end{document}